\documentclass[conference]{IEEEtran}
\IEEEoverridecommandlockouts
\usepackage{cite}
\usepackage{amsmath,amssymb,amsfonts}
\usepackage{algorithmic}
\usepackage{graphicx}
\usepackage{booktabs}
\usepackage{url}
\usepackage{color}
\usepackage{textcomp}
\usepackage{hyperref}
\usepackage{xcolor}
\def\BibTeX{{\rm B\kern-.05em{\sc i\kern-.025em b}\kern-.08em
    T\kern-.1667em\lower.7ex\hbox{E}\kern-.125emX}}
\begin{document}

\title{\textbf{Predictive Analysis of Tuberculosis Treatment Outcomes Using Machine Learning    : A Karnataka TB Data Study at a Scale}

\thanks{}
}

\author{
\IEEEauthorblockN{ SeshaSai Nath Chinagudaba}
\IEEEauthorblockA{\textit{Department of Mathematics and Computer Sciences} \\
\textit{Sri Sathya Sai Institute of Higher Learning}\\
Prasanthi Nilayam, Andhra Pradesh, INDIA \\
seshasainath.ch@gmail.com}
\and
\IEEEauthorblockN{ Dr. Darshan Gera}
\IEEEauthorblockA{\textit{Department of Mathematics and Computer Sciences} \\
\textit{Sri Sathya Sai Institute of Higher Learning}\\
Prasanthi Nilayam, Andhra Pradesh, INDIA \\
darshangera@sssihl.edu.in}
\and
\IEEEauthorblockN{ Dr. Krishna Kiran Vamsi Dasu\footnotemark[1]}
\IEEEauthorblockA{\textit{Department of Mathematics and Computer Sciences} \\
\textit{Sri Sathya Sai Institute of Higher Learning}\\
Prasanthi Nilayam, Andhra Pradesh, INDIA \\
dkkvamsi@sssihl.edu.in}
\and
\IEEEauthorblockN{ Dr. Uma Shankar S}
\IEEEauthorblockA{\textit{Divisional Head,  Epidemiology and Research} \\
\textit{National Tuberculosis Institute(NTI)}\\
Bengaluru, Karnataka, INDIA \\
ushankars.2017@gmail.com}
\and
\IEEEauthorblockN{ Dr. Kiran K}
\IEEEauthorblockA{\textit{NTEP Medical Consultant} \\
\textit{WHO}\\
Bengaluru, Karnataka, INDIA \\
kiranks24@gmail.com}
\and
\IEEEauthorblockN{ Dr. Anil Singarajpure}
\IEEEauthorblockA{\textit{State TB Officer} \\
Bengaluru, Karnataka, INDIA \\
dadranil@gmail.com}
\and
\IEEEauthorblockN{ Dr. Shivayogappa.U}
\IEEEauthorblockA{\textit{MBBS, D.Ortho} \\
\textit{Joint Director (TB), STO}\\
Bengaluru, Karnataka, INDIA \\
stoka@rntcp.org}

\and
\IEEEauthorblockN{ Dr. Somashekar N}
\IEEEauthorblockA{\textit{Director} \\
\textit{National Tuberculosis Institute(NTI)}\\
Bengaluru, Karnataka, INDIA \\
tbsoma@gmail.com}
\and
\IEEEauthorblockN{ Dr. Vineet Kumar Chadda}
\IEEEauthorblockA{\textit{EX-Advisor} \\
\textit{National Tuberculosis Institute(NTI)}\\
Bengaluru, Karnataka, INDIA \\
vineet2chadha@gmail.com}
\and
\IEEEauthorblockN{ Dr. Sharath B N.}
\IEEEauthorblockA{\textit{Professor of Community Medicine} \\
\textit{ESI Medical College}\\
Bengaluru, Karnataka, INDIA \\
sharath.burug@gmail.com}
}

\maketitle
\footnotetext[1]{Corresponding author}



\section*{\textbf{Abstract}}

Tuberculosis (TB) remains a global health threat, ranking among the leading causes of mortality worldwide. In this context, machine learning (ML) has emerged as a transformative force, providing innovative solutions to the complexities associated with TB treatment.This study explores how machine learning, especially with tabular data, can be used to predict Tuberculosis (TB) treatment outcomes more accurately. It transforms this prediction task into a binary classification problem, generating risk scores from patient data sourced from NIKSHAY, India's national TB control program, which includes over 500,000 patient records.\newline

Data preprocessing is a critical component of the study, and the model achieved an recall of 98\% and an AUC-ROC score of 0.95 on the validation set, which includes 20,000 patient records.We also explore the use of Natural Language Processing (NLP) for improved model learning. Our results, corroborated by various metrics and ablation studies, validate the effectiveness of our approach. The study concludes by discussing the potential ramifications of our research on TB eradication efforts and proposing potential avenues for future work. This study marks a significant stride in the battle against TB, showcasing the potential of machine learning in healthcare. \newline

\textbf{Keywords:} Machine Learning(ML), Tuberculosis (TB), Models, Binary Classification, Data Cleaning , Ensemble Learning, Performance Metrics, Evaluation.

\section{\textbf{Introduction}}
Tuberculosis (TB), caused by the bacterium \textit{Mycobacterium tuberculosis}, remains a persistent and significant global health threat, ranking among the leading causes of mortality worldwide. In 2023, India achieved a record in TB notifications with \textbf{25,37,235} reported TB cases. This includes cases reported in both the public sector, which totaled \textbf{16,99,119}, and the private sector, which was \textbf{8,38,161} \cite{india2023}. Despite being preventable and curable,TB continues to have a significant impact. Complex treatment procedures and the looming challenge of drug resistance pose significant obstacles to effective management.

The severity of TB is determined not only by its prevalence but also by the variability of treatment outcomes among patients. In Karnataka, there were \textbf{81,331} TB notifications in 2022 \cite{karnataka2022}. Furthermore, the emergence of drug resistance emphasizes the necessity for accurate and timely interventions to effectively combat the disease. The estimated tuberculosis prevalence-to-notification (P:N) ratio in Karnataka is \textbf{4.08}, which is significantly higher than the national average of \textbf{2.843} \cite{karnataka2022}. This means that for every reported case of  highly infectious disease, the actual number of cases in Karnataka is \textbf{4.083}.

In this context, machine learning (ML) has emerged as a transformative force, providing innovative solutions to the complexities associated with TB treatment. ML algorithms, such as random forest, support vector machines, and artificial neural networks, show the potential of extensive tabular data \cite{mlhealthcare}. This data-centric approach enables algorithms to analyze complex patterns within patient information, leading to accurate predictions of treatment outcomes \cite{tbml1, tbml2}.\newline

These ML techniques represent a paradigm shift in healthcare by helping physicians customize treatments based on individual patient profiles. Machine learning not only predicts treatment outcomes but also plays a crucial role in early detection, contact tracing, and epidemiological surveillance, contributing to more efficient public health interventions \cite{mlhealthcare}.\newline

Focused on the intersection of data science and healthcare, this study explores how machine learning, especially with tabular data, plays a crucial role in predicting TB treatment outcomes. By elucidating the symbiotic relationship between technology and medicine, this research aims to provide insights that can significantly influence the prognosis, management, and eventual control of this widespread global health crisis.

\section*{\textbf{Problem Statement}}
\textbf{Machine Learning for Trea   tment Outcome Prediction: } We utilize the  machine learning to predict the outcomes of a patient's treatment, which  interprets as a binary classification task. In this scenario, the target label $y_p$ is assigned 1 for a successful treatment and 0 otherwise. The features specific to each patient are represented as $x_p = \{x_{p1}, x_{p2}, ..., x_{pd}\}$. The model subsequently generates a risk score $\hat{y}_p = f (x_p)$, which corresponds to the probability of a patient experiencing a successful treatment outcome, where $f : \mathbb{R}^d \rightarrow [0, 1]$. We can assemble the features of $n$ patients into an $n \times d$ feature matrix $X$. The actual and predicted labels are denoted as $y$ and $\hat{y}$, respectively.\newline

\begin{figure*}[htbp]
\centerline{\includegraphics[scale=0.81]{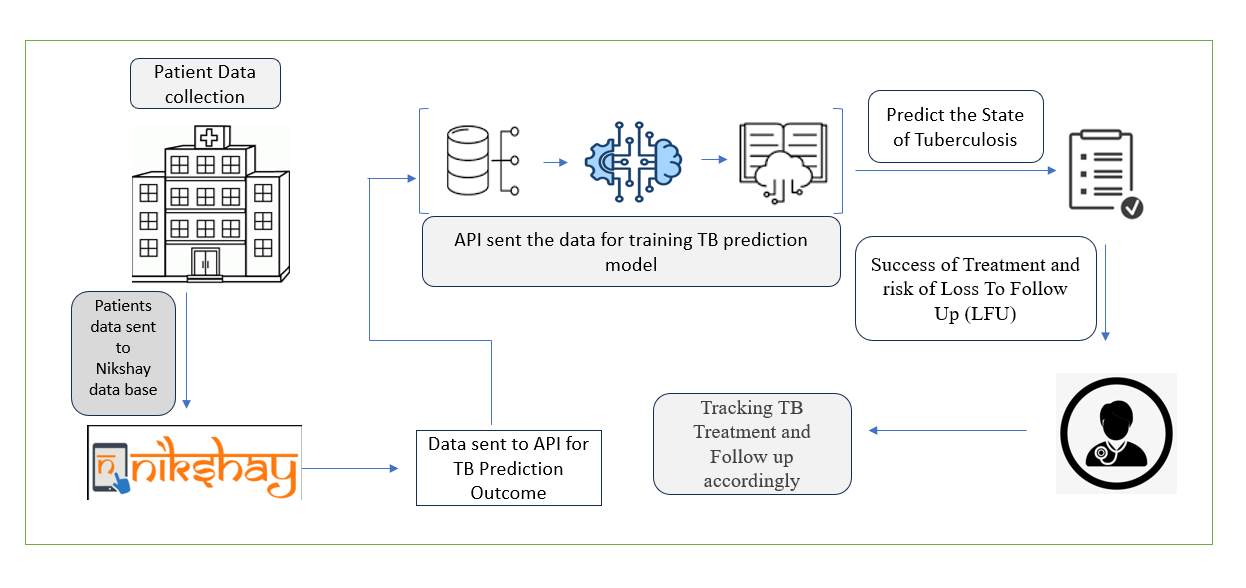}}
\caption{Process of TB Patient Treatment}
\label{PTPT}
\end{figure*}

\textbf{Early Stratification of TB Patients:}
Our approach employs machine learning methodologies to propose an early stratification of TB patients. This strategy enables the implementation of efficient, personalized interventions and care by transforming patient monitoring workflows and operational protocols.  This approach aids in mitigating the transmission of TB within each cohort such as Age, Gender, Diagnosis facility TBU and Type of case.  This strategy not only optimizes resource allocation but also ensures that patients receive the most appropriate care based on their individual risk profiles as shown in \textcolor{blue}{Figure \ref{PTPT}}. This could potentially revolutionize the way TB care is delivered, making it more efficient, effective, and patient-centric. Our research is a step towards a future where machine learning aids in healthcare, improving outcomes and saving lives.\newline

 We introduce a weighting scheme to the risk score calculation. With $w_i$ denoting the weight for the $i$-th feature, the risk score can be calculated as:

\begin{equation}
\hat{y}_p = f \left(\sum_{i=1}^{d} w_i x_{pi}\right)
\end{equation}

In our model, the predicted output, $\hat{y}_p$, is computed as a function $f$ of the weighted sum of the inputs, $\sum_{i=1}^{d} w_i x_{pi}$. The function $f$ serves as an activation function, introducing non-linearity into the model and enabling it to learn more complex patterns in the data. In this study, we have chosen $f$ to be a Rectified Linear Unit (ReLU), which is defined as $f(x) = \max(0, x)$.\newline

The weights can be learned during the training process, which allows the model to pay different attention to different features when predicting the treatment outcome. This adds an extra layer of complexity and adaptability to the model.



\section*{\textbf{Contributions}}

\subsection{\textbf{Framing the Challenge}}
We harness the power of machine learning to tackle a global public health issue that affects millions. Our model is crafted to allocate resources efficiently in a scenario where resources are limited. It's designed with real-world deployment considerations and constraints in mind to maximize its effectiveness. We use a unique evaluation method for model selection and split the data temporally (forward splitting) to accurately mimic the deployment scenario.

\subsection{\textbf{Enhancing the Model}}
Our dataset stands out due to its large size, infrequent target occurrence, a multitude of high-cardinality categorical attributes, and changes in data distribution due to cyclical shifts and the healthcare system's evolving focus towards other epidemics. In this context, we carry out an exhaustive search for the best combination of encoding and model, including advanced encodings for categorical variables, tabular deep learning models, and fully interpretable boosting algorithms. We also experiment with techniques like data augmentation, ensembling, interpretability, and algorithmic fairness to steer deployment and support public health authorities.

\subsection{\textbf{Assessment and Broad Applicability}}
We set aside six months of data for a passive evaluation to cover most out-of-distribution cases. Our models undergo rigorous evaluation across various cohorts of interest, including geographical, gender-based, State of TB, and Type of TB Unit health facility. We demonstrate strong generalization over time and across geographical locations, and we provide solid measures of predictive multiplicity \cite{marx2020} as evidence of the robustness of our model scores.

\subsection{\textbf{Potential for Positive Impact in Karnataka}}
The goal of our project is to roll out the solution systematically across all districts in Karnataka. This information will be leveraged by field staff, allowing for targeted use of health workers' capacity. As per the Union health ministry's recent estimate (2015), there are at least 217 positive cases of tuberculosis per 1 lakh people annually \cite{timesofindia2024}. In Karnataka, 101 cases are tracked per 1 lakh population as not all cases are officially recorded \cite{tbcindia2024}. \newline

For the patients who began treatment in the last six months of 2022, our machine learning solution could potentially make a significant difference in a large number of lives, compared to random targeting and the best rule-based baseline model. We are currently collaborating closely with the Karnataka state government for a pilot deployment, which could lead to a significant improvement in the state's TB treatment outcomes by effectively utilizing health workers’ bandwidth \cite{nhmkarnataka2024, thehindu2024}.\newline

\section{\textbf{Existing Approaches}}
In the realm of Tuberculosis (TB) research, various machine learning models have been employed to aid in the prediction and diagnosis of the disease.
\subsection*{\textbf{Approaches of TB in Machine Learning}}
A study by  Tiwari et al. reviewed the various causes and symptoms of tuberculosis and how machine learning techniques have been used for accurate and timely prediction and diagnosis over the past few years. They emphasized the role of machine learning in assisting physicians to diagnose and provide suitable treatment\cite{tiwari2019}. A recent academic study scrutinized the limitations of conventional methods used for diagnosing Tuberculosis (TB). It provided a comprehensive review of various machine learning algorithms and their significance in diagnosing TB\cite{unknown2024}. The study also delved into an array of deep learning strategies, which were integrated with other systems such as neuro-fuzzy logic, genetic algorithms, and artificial immune systems.\newline

A study published in Hindawi applied machine learning algorithms such as random forest, k-nearest neighbors, support vector machine, logistic regression, leaset absolute shrinkage and selection operator (LASSO), artificial neural networks (ANNs), and decision trees to analyze the case-control dataset for predicting Multidrug-Resistant Tuberculosis\cite{hindawi2024}. These studies exemplify the diverse applications of machine learning in TB research, highlighting the potential of these techniques in enhancing the prediction, diagnosis, and treatment of this global health challenge.\newline

\subsection*{\textbf{Approaches of TB in Deep Learning}}

In a comprehensive study, the authors explored the application of deep learning techniques for detecting Tuberculosis (TB) from chest radiographs \cite{olokooba2023}. They examined the most recent advancements in deep learning for screening pulmonary abnormalities associated with TB. Their systematic review, adhering to PRISMA guidelines, emphasized the potential of Computer-Aided Diagnostic (CAD) systems in addressing the challenges posed by the TB epidemic.\newline

In parallel, there has been a significant increase in research over the past decade, investigating machine learning techniques for analyzing chest X-ray (CXR) images to screen for cardiopulmonary abnormalities \cite{unknown2023}. This surge in interest aligns with the remarkable progress in deep learning, primarily based on convolutional neural networks (CNNs). These advancements have led to substantial contributions in the field of deep learning techniques for TB screening using CXR images.\newline

 This study explores the use of Deep Learning (DL) for predicting TB treatment outcomes from tabular data. Despite challenges such as large data requirements and lack of interpretability, DL’s potential for accuracy in predictions is being investigated.

\subsection*{\textbf{Approaches of TB in Ensembling ML}}
Ensemble learning, a powerful machine learning concept, has been extensively explored in the field of Tuberculosis (TB) research. A study by M.A. Ganaie et al. provided an extensive review of deep ensemble models, categorizing them into bagging, boosting, stacking, and other types\cite{ganaie2022}.
Another research proposed a robust framework for diabetes prediction, which is similar to TB prediction, where they employed different Machine Learning classifiers and proposed a weighted ensembling of these models\cite{diabetes2023}. Furthermore, ensemble algorithms have been applied in many practical applications to improve prediction accuracy, as highlighted in a review of ensemble learning algorithms used in remote sensing\cite{remote2022}. These studies underscore the potential of ensemble learning in enhancing the prediction and diagnosis of diseases like TB.

\section{\textbf{Data}}

In this study, we utilized data from tuberculosis patients obtained through the Nikshay system. Nikshay\footnote{\url{https://reports.nikshay.in/}}, a nationwide system for managing TB patients which established and maintained by the Central TB Division (CTD) of the Indian Government, the National Informatics Centre (NIC), and the World Health Organization (WHO) \cite{nikshay}. Furthermore, the support from the National Tuberculosis Institute (NTI) is of immense value. The NTI, an institution of the Indian Government, is committed to advanced tuberculosis research. It functions under the Directorate General of Health Services within the Ministry of Health and Family Welfare \cite{nti}. The NIKSHAY database contains tuberculosis patient data from Karnataka for 2021 and 2022, with approximately  \textbf{1,53,000} data points per record. The notification register and comorbidity register contain medical test results, comorbidities, and factors that impact the progression of TB. Key attributes in these registries can help predict tuberculosis treatment outcomes.\newline

Further details regarding both these registers include the following: \newline

\textbf{Notification Register:} This register contains data related to patient demographics, such as age, gender, geographical location, and socio-economic factors. It provides insights into the diagnosis process, the initiation of treatment, and the categorization of patients based on the severity of TB. Furthermore, it monitors the progression of TB treatment, including medication adherence and treatment outcomes.\newline

\textbf{Comorbidity Register:} This register is a key component that focuses on the additional health conditions and comorbidities common among TB patients. It includes comprehensive information on co-existing diseases, such as diabetes, HIV/AIDS, and other chronic illnesses. Moreover, it highlights habits and lifestyle factors that contribute to the worsening of TB and its comorbidities.\newline

After merging both notification and comordibity registers, a typical data point consists of the following attributes.\newline

\begin{enumerate}
    \item  \textit{Personal details}: Name, age, gender, address,  occupation. 

    \item  \textit{Medical history}: TB diagnosis details,  treatment  regimen, duration of  treatment. 

    \item  \textit{Test results}: Laboratory findings, imaging tests, drug susceptibility tests. 

    \item  \textit{Comorbidities}: Presence of other health conditions affecting TB patients.

    \item \textit{Lifestyle factors}: Habits and behaviors potentially impacting TB  progressionn (Key Population).\newline
\end{enumerate}

\begin{table}[h!]
  \centering
  \caption{Data Distribution }
  \label{tab:dd}
  \setlength{\tabcolsep}{4pt}
  \renewcommand{\arraystretch}{1.8}
  \begin{tabular}{ c c c c c } 
\hline
 & \textbf{Total} & \textbf{Cured (\%)} & \textbf{LFU (\%)} \\ \hline
\textbf{Training Data} & 119024 & 77.87 & 22.13 \\ 
\textbf{Evaluation Data} & 29757 & 77.85 & 22.15 \\ \hline
\end{tabular}

\end{table}

The dataset in Table \ref{tab:dd} exemplifies a case of imbalanced data, a common phenomenon in medical datasets. This imbalance is marked by a significant discrepancy in the number of instances between classes, such as patients who are cured versus those under Lost to Follow-Up (LFU) condition. The existence of imbalanced data can instigate a bias in the predictive model towards the majority class, undermining the model's predictive precision for the minority class. This is particularly detrimental in medical scenarios where the minority class often represents critical conditions that require accurate prediction. For instance, in the context of Tuberculosis (TB) treatment, the minority class could represent patients under LFU condition, a critical state that requires immediate medical attention and accurate prediction for effective intervention. Conversely, the majority class could represent patients who have been successfully cured. Therefore, addressing this imbalance is crucial for enhancing the model's performance in predicting both cured and LFU conditions accurately.\newline

The act of balancing the data is of paramount importance to ensure that the predictive model offers a reliable representation across all classes. A multitude of techniques, including oversampling of the minority class, undersampling of the majority class, or synthetic data generation methods like SMOTE, can be employed to attain data balance. A balanced dataset enables the model to learn from an equal number of instances from each class, thereby fostering a more generalized and robust model. This is particularly crucial in the context of predicting tuberculosis treatment outcomes, where each patient's data point holds substantial clinical significance.\newline

\subsection*{\textbf{Balancing Dataset}}

In the realm of medical data analysis, imbalanced datasets, where instances of one class greatly outnumber the other, are a common occurrence. This imbalance can lead to models that are biased towards the majority class. To mitigate this, techniques such as the Synthetic Minority Over-sampling Technique (SMOTE) and Random Over-sampling are employed.\newline

\textbf{SMOTE} is a technique that generates synthetic instances of the minority class. It achieves this by selecting two or more similar instances (based on a distance measure) from the minority class and altering an instance one attribute at a time by a random amount within the difference of the neighboring instances \cite{chawla2002smote}.\newline

\textbf{Random Over-sampling},  increases the representation of the minority class in the dataset by randomly duplicating instances from the minority class. This can be done in two ways: with replacement, where the same instance can be selected and added multiple times to the dataset; and without replacement, where each selected instance, once added, is not considered again \cite{he2008adasyn}. While these techniques can help balance the class distribution and improve the performance of machine learning models on imbalanced datasets, they should be used with caution as they can lead to overfitting while increasing the  minority class instances.

\section{\textbf{Data Prepossessing and Data splitting }}
\subsection{\textbf{Data Cleaning}}
This section outlines a systematic approach for cleaning the TB patient data, which primarily consists of categorical variables and a small number of numerical values. The cleaning process involves:\newline 

\begin{itemize}
    \item Segregating the required attributes.
    \item Addressing Data Discrepancies.
    \item Handling Missing Values.\newline
\end{itemize}
\newblock
\textbf{Attributes Segregation}: Identify and extract relevant attributes from the dataset, focusing on important attributes related to patient demographics, medical history, test results, comorbidities, and lifestyle factors.\newline

\textbf{Data Checking for Discrepancies}: Examine the dataset thoroughly for any discrepancies, such as misspelled words, inconsistent data entries, or unexpected values. Address any apparent errors to ensure the accuracy and integrity of the data.\newline

Based on discussion with NTI experts and ESIMC Resource members, we performed the following data cleaning to obtain the final TB patient data in order to achieve the proposed objective.\newblock \newline

\textbf{Addressing missing values:} If the missing values in a attributes exceed 15\%, consider removing the entire  attribute, as it may not contribute significantly to the analysis. For attributes with missing values below 15\%, each field should be addressed individually. \newline 

Here is an example of how to address missing values. In the dataset under study, we tackle missing values in a two-fold manner. For the ``Weight'' attribute, which has 4\% missing data, we first substitute these with the corresponding ``Age'' values, assuming no missing values in the ``Age'' attribute. Any remaining missing values are then filled using mean imputation. For the ``Basis of Diagnosis'' attribute, we cross-verify missing values with other data points. If the information is consistent and aligns with the diagnosis basis or test name, we fill in the missing values accordingly. This comprehensive approach ensures reliable data handling. \newline

Some of the methods used for  addressing data handling and
data cleaning are as follows.
\begin{itemize}
    \item Replacement of Values
    \item Strict Replacement
    \item Handling Missing Values
    \item Binary replacement for Target attribute.
\end{itemize}




\subsection{\textbf{Data Splitting}}
The model, which is trained on historical data related to Tuberculosis, is designed to predict future trends. This process involves using past data to understand the patterns and then applying this understanding to forecast future outcomes. The data is divided in a chronological manner. This means that the data is split based on the time of occurrence, maintaining the order in which the data was collected. This is crucial when dealing with time-series data like medical records, where the sequence of data points is important.\newline

The most recent six months of data are set aside for the final evaluation. This segment is known as the passive evaluation split, denoted as \[(y^{(ps)}, X^{(ps)}).\] This portion is not used for any aspect of model training or optimization. It's to test the model's performance after all the training and tuning processes are completed.

The remaining data, referred to as the modeling split \[(y^{(ms)}, X^{(ms)}),\] is further divided into training, validation, and test subsets. The division is done in a proportion of \textbf{70:15:15}. \newline

The training set, constituting 70\% of the modeling split, is used to train the model on the historical data. This is where the model learns the patterns in the data. The validation set, constituting 15\% of the modeling split, is used for hyperparameter tuning. Hyperparameters are parameters that are not learned from the data, but are set prior to the commencement of training. They guide the learning process and tuning them correctly can significantly improve the model's performance.\newline

The test set, constituting the remaining 15\% of the modeling split, is used to evaluate the model's performance and select the best model class. This helps in understanding how well the model is likely to perform on unseen data.\newline

Once the optimal model class is identified, it is trained on the entire modeling split and then subjected to its final evaluation on the passive evaluation split. The choice of the 70:15:15 division was empirically driven, aiming to strike a balance between having sufficient training data and enabling evaluation on a cohort basis. This ensures that the model is well-trained and also robustly validated and tested before being used for predictions.

\section{\textbf{Methodology}}
\subsection{\textbf{Encoding Technique}}

Encoding is a method of converting categorical variables into numerical values, making it easier to fit them into a machine learning model \cite{morita2024}. As the TB data has the most categorical attributes with the highest number of unique values, we need to ensure that the data is converted to numerical format. Since most models do not work with mixed data types (categorical and numerical), we need to encode the features into a real-valued vector space \cite{xing2024}.\newline

The encoding techniques here are used in two types. The one that uses the target attribute as information to encode the tabular data, such as Target Encoding, Ordinal Encoding, Leave-one-out Encoding, Catboost Encoding, and Odds-ratio Encoding, and some who do not use the target attribute, such as Similarity Encoding, Normalised-Count Encoding, and so on \cite{morita2024}.\newline

These encoded tables will be trained using high-depth XGBoost trees to achieve the best performance. The XGBoost trains the data much faster and delivers best performance compared to other machine learning methods \cite{xing2024}.\newline

A hybrid approach to this challenge involves the use of \textbf{Weight of Evidence (WoE)} and \textbf{Information Value (IV)}. WoE provides a powerful transformation of categorical variables into a format that is easily interpretable and effective for prediction. The IV is used to rank these variables in order of importance.\newline

The formulas for WoE and IV are given by:

\begin{equation}
WoE = \ln \left( \frac{\% \text{ of non-events}}{\% \text{ of events}} \right)
\end{equation}

\begin{equation}
IV = \sum \left( (\% \text{ of non-events} - \% \text{ of events}) \times WoE \right)
\end{equation}

After encoding, the data can be trained using XGBoost, a decision-tree-based ensemble Machine Learning algorithm that uses a gradient boosting framework. The objective function of XGBoost is given by:

\begin{equation}
\text{Obj}(\Theta) = \sum_{i=1}^n l(y_i, \hat{y}_i) + \sum_{k=1}^K \Omega(f_k)
\end{equation}

This approach provides a robust method for handling categorical variables in imbalanced datasets, delivering superior performance and allowing for the interpretation of the importance and effect of each variable on the prediction.

\subsection{\textbf{Metric }}

In the realm of healthcare, particularly in managing diseases like Tuberculosis (TB), machine learning solutions are pivotal. They aim to achieve high recall, identifying as many patients as possible who are likely to have unfavorable treatment outcomes, allowing for early intervention and better disease management \cite{mlhealthcare}. However, due to resource constraints and other responsibilities, healthcare workers can typically focus on a certain percentage (k\%) of patients for intensive monitoring and interventions \cite{resourceconstraints}. This necessitates a strategic approach where the focus is shifted to the top k\% of patients, ensuring that those most at risk are given priority.\newline

Given the low prevalence (approximately 3-4\%) of the positive class among TB patients \cite{tbprevalence}, standard classification evaluation metrics may not be suitable. Machine learning offers a more nuanced approach to disease management by employing custom metrics better suited to the specific needs and constraints of TB management \cite{mlcustommetrics}. These models optimize the prediction of treatment outcomes, highlighting the transformative potential of machine learning in tackling global health challenges.\newline

\subsubsection{\textbf{Recall@k}}

Recall@k is a performance metric for ranking problems, measuring the model’s effectiveness in ranking positive cases high. It’s calculated as the proportion of actual positive cases in the top k positions of the ranked list. This metric is crucial when only the top k items can be acted upon due to resource constraints. By maximizing Recall@k, we ensure the most positive cases are within the top k positions. This is particularly relevant in healthcare for effective treatment distribution.

\subsubsection{\textbf{Average Recall (10,40)}}
Average Recall (10,40) is a metric used in ranking problems, specifically in medical research. It calculates the proportion of actual positive cases within the top 10 and 40 positions of a ranked list. This metric is crucial when resources are limited, ensuring the majority of positive cases are captured within these positions, optimizing resource use and potentially improving patient outcomes.

\subsection{\textbf{Models}}
In our research, we explored a range of machine learning models, employing a variety of methodologies. This included both traditional models, such as Naive Bayes, Decision Trees, Random Forest, and \text{k-Nearest Neighbors (k-NN)}, and more advanced boosting machine learning models like \text{Gradient Boosting Machine (GBM)}, \text{XGBoost}, \text{LightGBM}, \text{and CatBoost}.\newline


Among the advanced models, GBM, XGBoost, LightGBM, and CatBoost have demonstrated significant advancements in predictive modeling. These models are recognized for their ability to handle diverse data types, their robustness against overfitting, and their high performance in terms of speed and accuracy.\newline

Upon rigorous analysis, we discovered that XGBoost and GBM models exhibited commendable performance. However, the LightGBM model was particularly noteworthy, outperforming all other models in terms of recall. In contrast, XGBoost and GBM demonstrated balanced results across all encoding techniques, indicating their robustness and versatility. Meanwhile, TabNet, a deep learning model, delivered average performance in our experiments.\newline

In addition to above models, we also explored ensemble methods in machine learning. Ensemble methods, which combine predictions from multiple models, often outperform single models. They provide a way to balance out the individual weaknesses of a set of models. Through techniques like bagging, boosting, and stacking, ensemble methods can improve the predictive performance and robustness of machine learning systems. In our research, ensemble methods proved to be highly effective, often delivering the best performance among all methods tested.\newline

One more ensembling hybrid XGBoost-Light GBM model, the final prediction $P_{\text{final}}$ is a weighted average of the XGBoost and LightGBM predictions, $P_{\text{xgb}}$ and $P_{\text{lgbm}}$:

\begin{equation}
P_{\text{final}} = w_{\text{xgb}} \cdot P_{\text{xgb}} + w_{\text{lgbm}} \cdot P_{\text{lgbm}}
\end{equation}

The loss function $L$ is a weighted sum of the individual models' loss functions:

\begin{equation}
L(y, P_{\text{final}}) = w_{\text{xgb}} \cdot L(y, P_{\text{xgb}}) + w_{\text{lgbm}} \cdot L(y, P_{\text{lgbm}})
\end{equation}

Weights $w_{\text{xgb}}$ and $w_{\text{lgbm}}$ are determined by the performance of the individual models. This approach combines the strengths of both models for improved performance.

\subsection{\textbf{HyperParameter tuning}}

\subsubsection{Traditional methods}
Hyperparameter tuning refers to finding the best hyperparameters, which determine how a machine learning model learns. Conventional techniques involve grid search, which is a comprehensive search through a manually chosen part of all possible hyperparameter values to enhance the performance of the model.\newline

\subsubsection{Hyperopt Method}
Hyperopt performs hyperparameter optimization which utilizes Bayesian optimization and the Tree of Parzen Estimators algorithm (TPE) to efficiently search through large hyperparameter spaces. The purpose is to identify the optimal set of parameters and supports conditional dependencies between hyperparameters during the optimization process.\newline 

\subsubsection*{\textbf{Selection of ML Model}}
We perform hyperparameter tuning, model selection, and encoder selection using the evaluation metric 
AvRecall (10, 40) as the loss function $L$. This is instead of the {}traditional training loss. The entire process can be described as follows:\newline

\textit{Model Selection}: In the context of predicting tuberculosis (TB) treatment outcomes, a binary classification problem, we select the optimal model $f^*$ from a set of all models $F$. This is achieved by minimizing the loss function $L$, which could be a binary cross-entropy loss function, over the hyperparameters $\lambda$ of each model $f$ and the best encoding $E^*$, with a regularization term $R(f)$ that could be a measure of model complexity, such as the number of parameters in the model $f$.

\begin{equation}
f^* = \arg\min_{f \in F} \min_{\lambda \in \Lambda_f} [L(y, f(X; \lambda, E^*)) + \alpha R(f)]
\end{equation}

\textit{Encoder Selection:} We select the optimal encoding $E^*$ from a set of all encodings $E$. This is achieved by minimizing the loss function $L$ over the hyperparameters $\lambda$ of the XGBoost model $fxgb$, with a regularization term $R(E)$ that could be a measure of the complexity of the encoder $E$, such as the dimensionality of the encoded data.

\begin{equation}
E^* = \arg\min_{E \in E} \min_{\lambda \in \Lambda_{fxgb}} [L(y, fxgb(X; \lambda, E)) + \beta R(E)]
\end{equation}

In both equations, $\alpha$ and $\beta$ are hyperparameters that control the strength of the regularization. The choice of $R(f)$ and $R(E)$ depends on the specific problem and the set of models or encoders.\newline

\textbf{Encoder Performance :}
The performance of models is intrinsically linked to the encoding of categorical variables. Hence, the model $f(x)$ is adapted to $f(x; \lambda, E)$, where $\lambda$ and $E$ symbolize model hyperparameters and encoding respectively.\newline

{\textbf{Optimization Strategy}:}
Under ideal circumstances, both $F$ and $E$ would be optimized simultaneously. However, due to the extensive dataset size and the multitude of potential combinations, this is not practically feasible. Therefore, we first search for the optimal encoder using the XGBoost model, which has exhibited robust performance on tabular data. Additionally, we manually explore the hyperparameters of the encoders.\newline

The validation subset $X(ms,val)$ within the modeling split is used for tuning hyperparameters, while the corresponding test subset $X(ms,test)$ is utilized for the search of the encoder and model. Once the optimal model $f^*$ and its corresponding best hyperparameter set $\lambda^*$ are identified, the model is retrained on the entire $X(ms)$ set. For the final prediction, the best model is applied to the passive evaluation set or on various cohorts.

\begin{equation}
y^{\text{(ps)}} = f^{\text{*(ms)}}(x^{\text{(ps)}}; \lambda_{f}^*, E^*)
\end{equation}

\section{\textbf{Results with Evaluations }}
Data cleaning was the initial step, followed by the application of encoding techniques. High-depth XGBoost trees were then utilized. The best model, along with the best model selection with hyperparameter tuning, selected based on recall, will be discussed in further sections.\newline

\subsection*{\textbf{Modeling split Evaluation}}
Our study involved a dataset with 98\% categorical features, including several high-cardinality ones. The model performance was significantly influenced by encoding techniques, with similarity count and catboost encoding being the most effective.\newline

We evaluated models using these encodings, focusing on AvRecall (10, 40) and Recall@20. The XGBoost model excelled, achieving a recall rate of ~99\% with hypertuned, closely followed by LightGBM and GBM . \newline

Considering average recall, XGBoost outperformed all other models. An ensemble of the top 5 models slightly improved the metrics over XGBoost, possibly due to the low output correlation between the Gradient Boosting Machine (GBM) and other models. Our solution demonstrated robustness, with low predictive multiplicity across models.\newline

\subsubsection{Encoding Results}
The various categorical encoding techniques used in machine learning define several custom encoder classes, such as \texttt{ProbabilityRatioEncoder}, \texttt{LogOddsRatioEncoder}, \texttt{OddsRatioEncoder}, and \texttt{SimilarityCountEncoder} and so on. Each class has \texttt{fit} and \texttt{transform} methods for training the encoder on the data and then transforming the data. The code also includes an evaluation of these encoding techniques using an XGBoost classifier model. The model is trained on the encoded data and then used to make predictions. The best performing encoding techniques in terms of  Precision, Recall, and F1 Score were Cat Boost Encoding, Similarity Count Encoding, and WOE\_IV Encoding. These techniques consistently achieved high scores across all metrics. Please see  \textcolor{blue}{Table \ref{tab:encoding}}
 for more details.

\begin{table}[h!]
  \centering
  \caption{Evaluation Of Different Encoding Techniques}
  \label{tab:encoding}
  \setlength{\tabcolsep}{4pt}
  \renewcommand{\arraystretch}{1.8}
  \begin{tabular}{ c c c c c } 
    \hline
    \textbf{Technique} & \textbf{Accuracy} & \textbf{Precision} & \textbf{Recall} & \textbf{F1 Score} \\ 
    \hline
    Target Encoding & 0.87 & 0.88 & 0.97 & 0.92 \\ 
    Normalized Count Encoding & 0.88 & 0.88 & 0.98 & 0.93 \\ 
    \textbf{CatBoost Encoding} & \textbf{0.88} & \textbf{0.88} & \textbf{0.98 }& \textbf{0.92} \\ 
    Ordinal Encoding & 0.88 & 0.88 & 0.98 & 0.93 \\ 
    Leave-One-Out Encoding & 0.87 & 0.88 & 0.97 & 0.92 \\ 
    Probability Ratio Encoding & 0.87 & 0.88 & 0.97 & 0.92 \\ 
    Log-Odds Ratio Encoding & 0.78 & 0.78 & 1.00 & 0.88 \\ 
    Odds Ratio Encoding & 0.87 & 0.88 & 0.97 & 0.92 \\ 
    \textbf{Similarity Count Encoding} & \textbf{0.88} & \textbf{0.88} & \textbf{0.98} & \textbf{0.93} \\ 
    TabNet & 0.88 & 0.87 & 0.97 & 0.93 \\
    W\textbf{OE\_IV Encoding}  & \textbf{0.89} & \textbf{0.89} & \textbf{0.98} & \textbf{0.93} \\
    
    \hline
  \end{tabular}
\end{table}

\begin{table}[h!]
\caption{Evaluation Of Encoding Techniques Adopted Metric Results}
\begin{center}

{ 
\setlength{\tabcolsep}{10pt}
\renewcommand{\arraystretch}{1.5}
\begin{tabular}{ c c c } 
 \hline
 \textbf{Technique} & \textbf{Recall@20} & \textbf{AvRecall(10, 40)} \\ 
 \hline
 Target Encoding & 0.917 & 0.862 \\ 
 Normalized Count Encoding & 0.920 & 0.864 \\
 \textbf{CatBoost Encoding} & \textbf{0.945} & \textbf{0.873} \\
 Gap Encoding & 0.917 & 0.862 \\ 
 MinHash Encoding & 0.917 & 0.862 \\ 
 Probability Ratio Encoding & 0.939 & 0.890 \\ 
 Log-Odds Ratio Encoding & 1.000 & 1.000 \\ 
 Odds Ratio Encoding & 0.939 & 0.890 \\ 
 \textbf{Similarity Count Encoding} & \textbf{0.939} & \textbf{0.862} \\ 
 TabNet & 0.911 & 0.862 \\ 
 \textbf{WOE\_IV Encoding} & \textbf{0.932} & \textbf{0.875} \\
 \hline
\end{tabular}
} 
\end{center}

\end{table}

\newpage
\subsubsection{Model Results}
 CatBoost encoding, with conjunction with various machine learning models, is first fitted and transformed on the training data, and then applied to the test data. A variety of models, including XGBoost, LightGBM, CatBoost, DecisionTree, RandomForest, GradientBoosting, and KNN, are then trained using the encoded data.\newline
 Each model's performance is evaluated based on accuracy, precision, recall, and F1 score metrics. The XGBoost LightGbm works better over other models. The results were shown in \textcolor{blue}{Table \ref{tab:ml}} \newline 
 
\begin{table}[h!]
\caption{Evaluation Of Different Machine Learning models}
\label{tab:ml}
\centering
{ 
\setlength{\tabcolsep}{4pt}
\renewcommand{\arraystretch}{1.8}
\begin{tabular}{ c c c c c } 
 \hline
 \textbf{Model} & \textbf{Accuracy} & \textbf{Precision} & \textbf{Recall} & \textbf{F1 Score} \\ 
 \hline
 \textbf{XG Boost} & \textbf{0.879} & \textbf{0.874} & \textbf{0.987} & \textbf{0.927} \\ 

\textbf{ Light GBM} & \textbf{0.882} & \textbf{0.875} & \textbf{0.990} & \textbf{0.929} \\ 

 Cat Boost & 0.882 & 0.874 & 0.986 & 0.929 \\ 

 Decision Tree & 0.808 & 0.875 & 0.879 & 0.877 \\ 

 Random Forest & 0.881 & 0.874 & 0.990 & 0.928 \\ 

 Gradient Boosting & 0.880 & 0.873 & 0.991 & 0.928 \\ 
 
 KNN & 0.867 & 0.874 & 0.968 & 0.919 \\ 
Avg. Ensemble & 0.871 & 0.874 & 0.978 & 0.911 \\
 \hline
\end{tabular}
} 

\end{table}
\newblock
 The models are not hyper-tuned in this process. Hyper-tuning could potentially improve the performance of the models by optimizing their parameters. Comparing the results before and after hyper-tuning would provide valuable insights into the effectiveness of the encoding technique and the models used. The results are shown below in \textcolor{blue}{Table \ref{tab:hyp}}.\newline

\begin{table}[h!]
\caption{ Machine Learning Models with Metric (Recall)}
\label{tab:avg}
\begin{center}
{ 
\setlength{\tabcolsep}{7pt}
\renewcommand{\arraystretch}{1.5}
\begin{tabular}{ c c c c } 
 \hline
 \textbf{Technique} & \textbf{Recall@20} & \textbf{AvRecall(10, 40)} \\ 
 \hline
 XG Boost & 0.982 & 0.977 \\ 
 Light GBM & 0.981 & 0.969 \\ 
 Cat Boost & 0.962 & 0.927 \\
 Decision Tree & 1.000 & 0.973 \\ 
 Balanced Random Forest & 0.999 & 0.999 \\ 
 Gradient Boosting & 0.957 & 0.931 \\ 
 KNN & 0.998 & 0.996 \\ 
Avg. Ensemble & 0.988 & 0.973 \\ 
 \hline
\end{tabular}
} 
\end{center}
\end{table}

\begin{table}[h!]
\caption{Hypertuned Machine Learning Models}
\label{tab:hyp}
\begin{center}
    
{ 
\setlength{\tabcolsep}{4pt}
\renewcommand{\arraystretch}{1.8}
\begin{tabular}{ c c c c c } 
 \hline
 \textbf{Model} & \textbf{Accuracy} & \textbf{Precision} & \textbf{Recall} & \textbf{F1 Score} \\ 
 \hline
 \textbf{XG Boost} & \textbf{0.879} & \textbf{0.867} & \textbf{0.997} & \textbf{0.928} \\
 
 \textbf{Light GBM} & \textbf{0.877} & \textbf{0.865} & \textbf{0.998} & \textbf{0.927} \\ 
 
 Cat Boost & 0.882 & 0.876 & 0.988 & 0.929 \\ 
 
 Decision Tree & 0.879 & 0.876 & 0.984 & 0.927 \\ 
 
 Random Forest & 0.880 & 0.871 & 0.993 & 0.928 \\ 
 
 Gradient Boosting & 0.876 & 0.863 & 0.999 & 0.926 \\ 
 
 KNN & 0.872 & 0.873 & 0.977 & 0.922 \\ 
Avg. Ensemble & 0.869 & 0.883 & 0.991 & 0.922 \\ 
 \hline
\end{tabular}
}
\end{center}
\end{table}

\newblock
\newpage
\subsection*{\textbf{Performance Over Passive Evaluation split Evaluation}}

\textbf{The research focused on using machine learning techniques to predict adherence to Tuberculosis (TB) treatment protocols.} The effectiveness of these techniques was gauged using two metrics: \textcolor{red}{Recall@20} and \textcolor{red}{AvRecall(10,40)}. Different encoding methods for handling categorical data were compared in conjunction with the \textcolor{blue}{XGBoost classifier}. The \textcolor{green}{Similarity Count Encoder} emerged as the most effective method, achieving a Recall@20 score of 0.973 and an AvRecall(10,40) score of 0.95. Reference to these results were shown in \textcolor{blue}{Table \ref{tab:avg}}\newline

Several machine learning models were evaluated, with the \textcolor{blue}{Average ensemble model} outperforming the others, achieving a Recall@20 score of 0.988 and an AvRecall(10,40) score of 0.973. When applied to a passive evaluation holdout set, the XGBoost model achieved a Recall@20 of 0.982 and AvRecall(10,40) of 0.977 in \textcolor{blue}{Table \ref{tab:avg}}. The average ensemble model performed slightly better, with a Recall@20 of 0.988 and AvRecall(10,40) of 0.973.\newline

Assuming a patient targeting rate of 20\%, the machine learning models could significantly enhance the treatment outcomes for TB patients. Specifically, the models could potentially assist approximately \textcolor{red}{5900 patients} out of \textcolor{red}{29757 patients} in adhering to their treatment regimen over a passive evaluation split period. The models demonstrated robust generalization capabilities, with no significant reduction in performance by month, suggesting that models trained on earlier data can effectively predict patient adherence in later months.

\subsection*{\textbf{Balanced dataset results}}

\subsubsection{\textbf{SMOTE}}
To prevent model bias towards the majority class in imbalanced datasets, various encoding techniques and Synthetic Minority Over-sampling Technique (SMOTE) are used. Techniques such as Target Encoding, Normalized Count Encoding, CatBoost Encoding, Ordinal Encoding, Log Odds Ratio Encoding, and Odds Ratio Encoding transform categorical variables. Each technique influences the model’s performance differently.\newline

SMOTE is then applied to the training set. It works by creating synthetic samples from the minority class, which helps balance the class distribution. This balanced dataset is then used to train an XGBoost model. The performance is assessed using accuracy, precision, recall, and F1 score. These metrics offer a holistic view of the model's effectiveness for both positive and negative classes, ensuring balance and impartiality with the following results below in \textcolor{blue}{Table \ref{tab:smote}}.\newline

\newblock

\begin{table}[h]
\centering
\caption{Model Performance Metrics with SMOTE Techniques}
\label{tab:smote}
\setlength{\tabcolsep}{6pt}
\renewcommand{\arraystretch}{1.5}
\begin{tabular}{lllll}
\hline
\textbf{Model} & \textbf{Accuracy} & \textbf{Precision} & \textbf{Recall} & \textbf{F1 Score} \\ \hline
\textbf{XG Boost} & \textbf{0.877} & \textbf{0.878} & \textbf{0.979} & \textbf{0.925} \\ 
Light GBM & 0.879 & 0.880 & 0.977 & 0.926 \\ 
Cat Boost & 0.882 & 0.876 & 0.988 & 0.928 \\ 
Decision Tree & 0.729 & 0.874 & 0.762 & 0.814 \\ 
Random Forest & 0.875 & 0.880 & 0.972 & 0.924 \\ 
Gradient Boosting & 0.878 & 0.881 & 0.976 & 0.926 \\ 
KNN & 0.733 & 0.889 & 0.751 & 0.814 \\
Avg. Ensemble  & 0.865 & 0.851 & 0.961 & 0.911 \\ \hline
\end{tabular}

\end{table}

\subsubsection{\textbf{Random Over-Sampling with Replacement}}
To address the issue of class imbalance, an alternative approach is to use Random Over-Sampling with Replacement. By duplicating examples from the minority class in the training dataset, and balances the class distribution.\newline

Similar to SMOTE, various encoding techniques such as Target Encoding, Normalized Count Encoding, CatBoost Encoding, Ordinal Encoding, Log Odds Ratio Encoding, and Odds Ratio Encoding are used to transform categorical variables. Each of these techniques influences the model's performance in unique ways.\newline

The Random Over-Sampling with Replacement is applied to the training set, this balanced dataset is then used to train an XGBoost model. The model's performance is assessed using accuracy, precision, recall, and F1 score. These metrics provide a comprehensive view of the model's effectiveness for both positive and negative classes, ensuring a balanced and impartial evaluation.\newline

This approach, like the one using SMOTE, allows for a fair comparison of model performance when dealing with imbalanced datasets. The results obtained in \textcolor{blue}{Table \ref{tab:rand}} would offer valuable insights into the effectiveness of Random Over-Sampling with Replacement in comparison to other resampling techniques.\newline

\begin{table}[h]
\centering
\caption{Model Performance Metrics with Random Over-Sampling}
\label{tab:rand}
\setlength{\tabcolsep}{6pt}
\renewcommand{\arraystretch}{1.5}
\begin{tabular}{lllll}
\hline
\textbf{Model} & \textbf{Accuracy} & \textbf{Precision} & \textbf{Recall} & \textbf{F1 Score} \\\hline
\textbf{XG Boost} & \textbf{0.865} & \textbf{0.881} & \textbf{0.955} & \textbf{0.916} \\
Light GBM & 0.869 & 0.888 & 0.952 & 0.919 \\
Cat Boost & 0.859 & 0.890 & 0.935 & 0.912 \\
{\textbf{Balanced Random Forest}} & \textbf{0.880} & \textbf{0.878} & \textbf{0.982} & \textbf{0.927} \\
Decision Tree & 0.813 & 0.877 & 0.883 & 0.880 \\
Random Forest & 0.879 & 0.877 & 0.983 & 0.927 \\
Gradient Boosting & 0.863 & 0.890 & 0.941 & 0.914 \\
KNN & 0.739 & 0.887 & 0.762 & 0.820 \\
Avg. Ensemble  & 0.839 & 0.871 & 0.941 & 0.890 \\\hline
\end{tabular}

\end{table}

\subsection*{\textbf{Interpretability}}

\subsubsection*{SHAP Analysis}
\begin{figure}[htbp]
 \centerline{\includegraphics[scale=0.3]{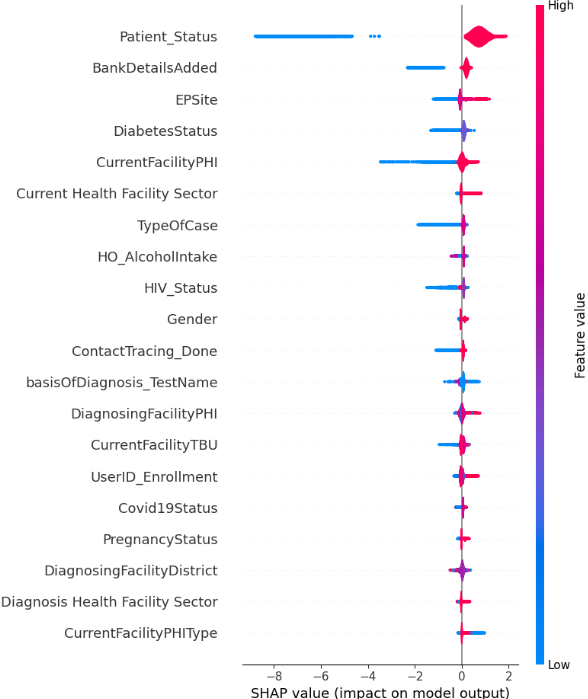}}
\caption{Shap Analysis}
\label{shap}
\end{figure}

SHapley Additive exPlanations (SHAP) analysis is a powerful tool for interpreting machine learning models. In the provided SHAP analysis graph, each variable such as "Patient Status," "BankloanBalance," "DiabetesStatus," and others, is plotted against their SHAP values. These values quantify the impact of each variable on the model's output. The graph provides a visual interpretation of how each variable influences the predictions of the model. \newline
From the \textcolor{blue}{Figure \ref{shap}}
, it appears that features like ``Patient\_Status,'' ``BankDetailsAdded,'' and ``EPSite'' have a high positive impact on the model's output as indicated by their position towards the right side of the scale. On the other hand, features like ``CurrentFacilityPHIType,'' ``Diagnosis Health Facility Sector,'' and others have a negative impact as they are positioned towards the left side of the scale.

This means that changes in features like ``Patient\_Status,'' ``BankDetailsAdded,'' and ``EPSite'' are likely to increase the model's output, while changes in features like ``CurrentFacilityPHIType,'' ``Diagnosis Health Facility Sector,'' and others are likely to decrease the model's output.

\subsubsection*{LIME}

\begin{figure}[htbp]
\centerline{\includegraphics[scale=0.3]{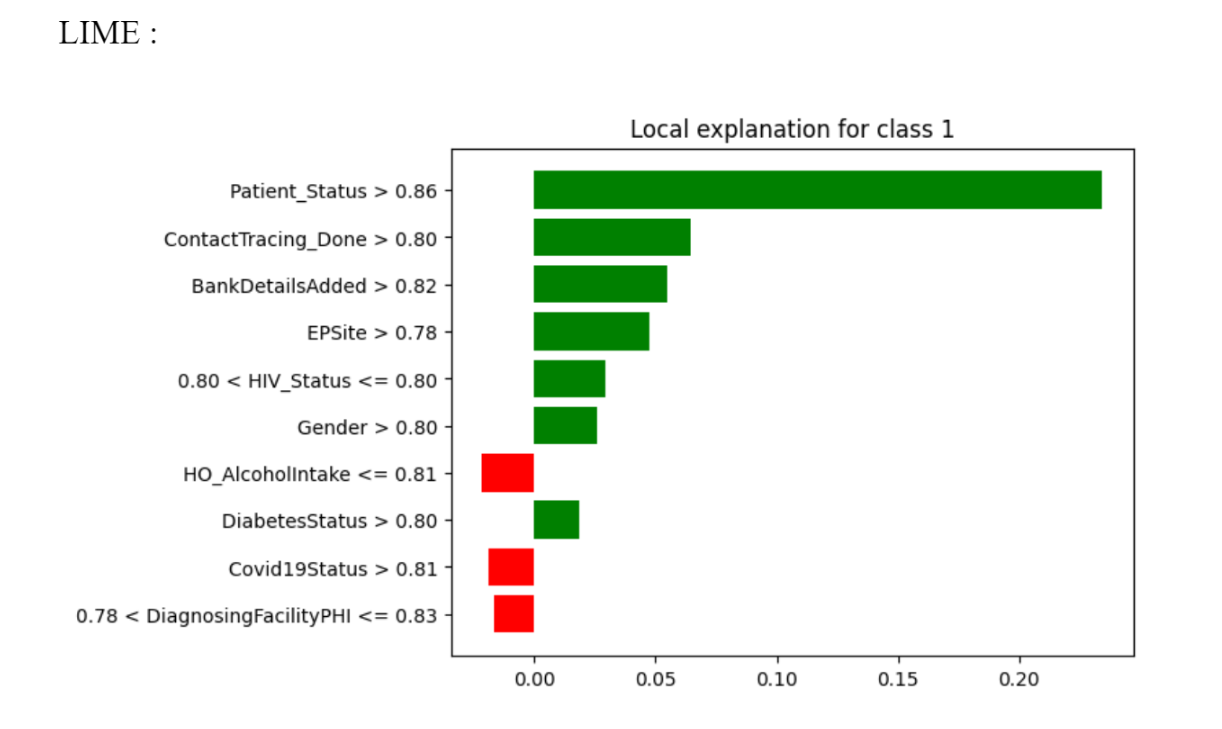}}
\caption{LIME Analysis}
\label{rangeq}
\end{figure}

The bar chart, titled "LIME: Local explanation for class 1 (Positive)", is a visual representation of a machine learning model's decision-making process. It signifies the importance of various factors in predicting class 1, which we can also refer to as "Positive". Factors such as Patient Status, ContactTracing Done, BankDetailsAdded, EPSite, HIV Status, Gender, and DiabetesStatus positively influence the prediction, while HQ AlcoholIntake and DiagnosingFacilityPHI negatively influence it. The numerical values associated with each factor indicate their significance in the model's prediction shown in \textcolor{blue}{Figure \ref{rangeq}}
. This analysis provides valuable insights into the model's behavior, contributing to its interpretability and transparency. It's an essential tool for understanding and validating the model's predictions, especially in critical applications where model decisions need to be explainable.\newline

\subsubsection*{Recall@k vs k}

The provided image is a graphical representation comparing the performance of seven distinct machine learning models: XGBoost, LightGBM, CatBoost, Decision Tree, Random Forest, Gradient Boosting, and KNN. The performance is evaluated based on the metric 'Recall@k', which is plotted against 'k'.\newline

\begin{figure}[htbp]
\centerline{\includegraphics[scale=0.3]{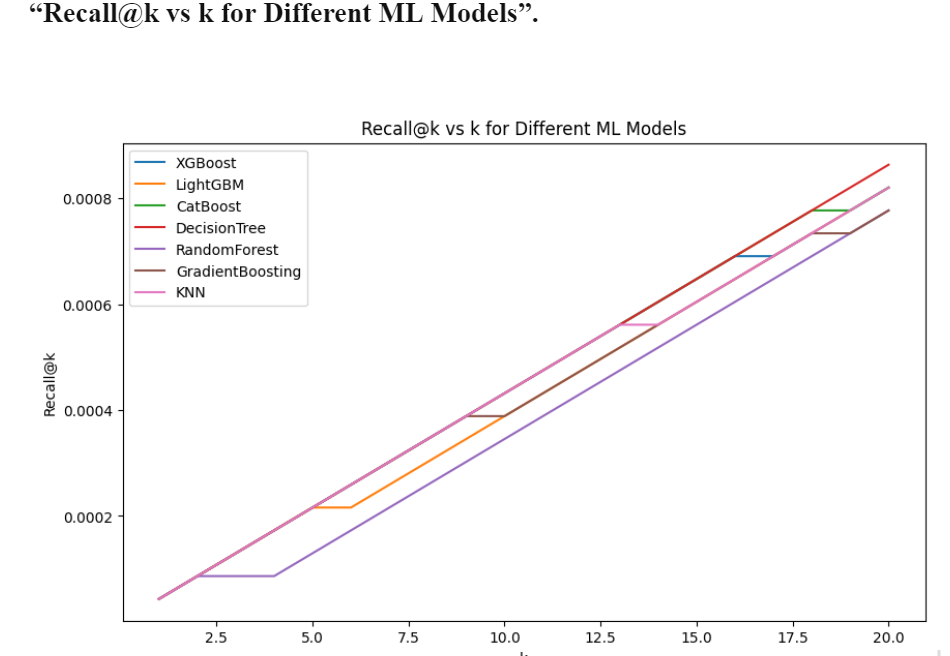}}
\caption{Recall@k vs k for different models}
\label{rec}
\end{figure}

From the \textcolor{blue}{Figure \ref{rec}}
, it is observed that as 'k' increases, the 'Recall@k' also increases for all the models. This implies that when we consider more predictions (increase 'k'), we are more likely to capture the relevant ones (increase Recall@k). \newline

However, it is important to note that while increasing 'k' does improve Recall@k, it may also bring in more irrelevant predictions. Therefore, the choice of 'k' often involves a trade-off between recall and precision.\newline


\newblock
\subsubsection*{ROC Curve}

The Receiver Operating Characteristic (ROC) is a performance measurement for binary classification problems in machine learning, which plots the True Positive Rate (TPR) against the False Positive Rate (FPR) at various threshold settings. The Area Under the Curve (AUC) is an aggregated measure of performance of a model across all possible classification thresholds. 

\begin{figure}[htbp]
\centerline{\includegraphics[scale=0.3]{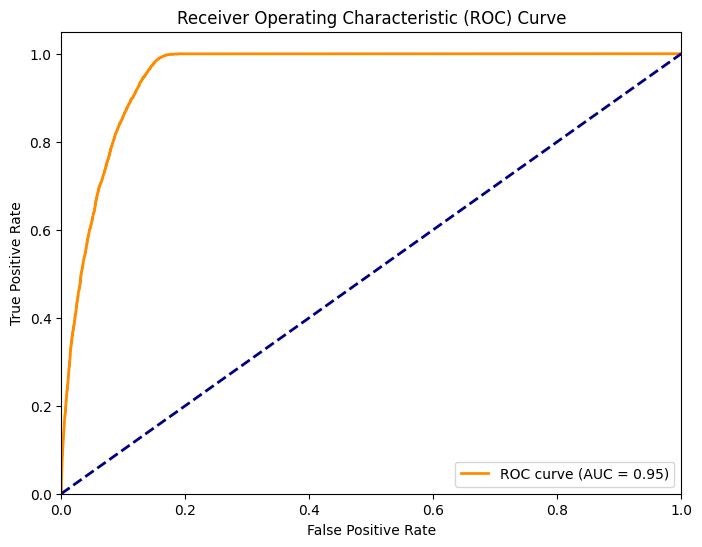}}
\caption{AUC-ROC Curve}
\label{roc}
\end{figure}

The AUC is 0.95 in \textcolor{blue}{Figure \ref{roc}}
, which is very close to 1. This indicates that the model has a high measure of separability. which classifies positive and negative classes very well defines that model is excellent at distinguishing between the two classes.The ROC curve provides a visual representation of the trade-off between the TPR and FPR for every possible cut-off, while the AUC gives a single numeric metric to compare models. The high AUC score of 0.95 in this case indicates an excellent model performance. 

\subsubsection*{Precision-Recall curve}

\begin{figure}[htbp]
\centerline{\includegraphics[scale=0.4]{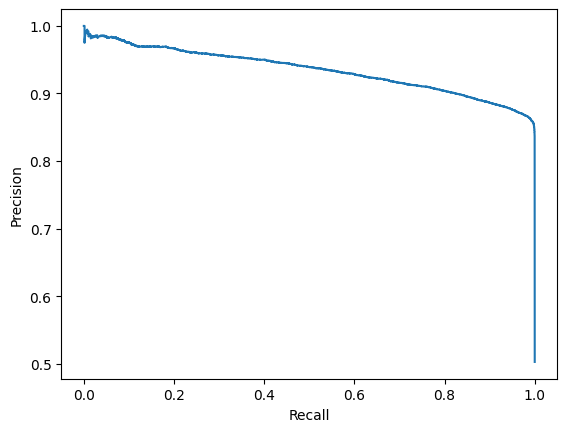}}
\caption{Precision-Recall curve}
\label{prec}
\end{figure}

 Precision-Recall curve, a statistical tool utilized in machine learning to assess the performance of a classification model. The curve \textcolor{blue}{Figure \ref{prec}}
is plotted with 'Recall' on the X-axis and 'Precision' on the Y-axis, both ranging from 0 to 1. The precision-recall curve, which commences at a high precision with low recall, maintaining this high precision until around a recall of 0.8, where it experiences a sharp decline. This curve is instrumental in comprehending the trade-off between precision and recall for varying threshold settings in a classification model. A model exhibiting perfect precision and recall would result in a curve reaching the top right corner of the plot, indicative of a high-quality classifier. This graphical representation serves as a critical measure of a model's classification performance, particularly in scenarios where classes are imbalanced.
 
\subsection*{\textbf{Advancing Outcomes for Lower Performing Cohorts}}

Our model has shown strong results across a variety of cohorts, but there is a discernible difference in outcomes when we look at age groups, TB units, and genders. To address this, we are considering two strategic solution.\newline

\subsubsection*{Data Expansion}
We are considering expanding our training data by adding more instances from the lower performing cohorts. This is akin to giving more weight to patients from these cohorts in the loss function.\newline

\subsubsection*{Ensuring Algorithmic Balance}
It's important to maintain balance in algorithms, especially in healthcare AI applications. We are looking into a post-hoc balance method that modifies model scores to ensure our metric is balanced across key cohorts such as age groups, TB units, and gender.\newline

These strategies have shown significant improvements in the performance of lower performing cohorts, particularly those with very low scores. The impact on higher-scoring cohorts has been minimal, and there has been a reduction in overall disparity. For example, TB units with a Recall@20 less than 0.7 have shown an 60\% increase in mean Recall@20 with data expansion. For gender, the balance method has also resulted in a slight improvement for the Gender cohort.\newline

 \section{\textbf{Strategic Implementation of Pilot Studies}}
We are in the process of planning pilot studies in various regions within Karnataka. We outline some measures taken to ensure responsible implementation and reserve a comprehensive description for future work.\newline

\subsection*{\textbf{Patient Safety:}} This device can be used as a   care, for high-risk patients receiving additional attention. We recommend not withholding care from patients who are already receiving it (via current interventions, e.g., therapy) and we encourage staff to exercise their discretion as needed. We clearly convey that our tool is designed to suppliment and enhance, existing practices. We also plan to conduct qualitative patient interviews and ground-truthing to validate treatment outcomes.\newline

\section{\textbf{Conclusions}}

This research has made significant strides in the field of machine learning, particularly in the development of predictive models for TB treatment outcomes. These models have shown remarkable effectiveness, especially for TB patient data. This is a crucial advancement, given the profound impact it could have on public health, particularly in regions like India where the burden of TB is high.\newline

 The study employed an ensembled hybrid learning approach, combining different model types to enhance overall performance. This approach, coupled with the use of Weight of Evidence (WOE) and Information Value (IV) encoding for categorical variables, resulted in robust and accurate predictions. Furthermore, the research explored the use of end-to-end learnable embeddings and enhancements to the ensembled hybrid learning approach. It also considered the application of Natural Language Processing (NLP) for improved model learning. These advancements are expected to be a significant asset for healthcare professionals once deployed.\newline

The study underscores the potential of machine learning in improving TB treatment outcomes globally. The insights gained are not only valuable for similar challenges but also highlight the broader applicability of machine learning in healthcare. The successful deployment of these models could mark a significant milestone in the fight against TB, demonstrating the transformative potential of machine learning in healthcare. This research, therefore, serves as a beacon, guiding future efforts in the application of machine learning for improved health outcomes.

\section{\textbf{Acknowledgment}}

This research project, focusing on Tuberculosis, has been made possible due to the substantial support from the Karnataka state Government Tuberculosis Division. Their commitment and the contributions which have enriched the project and contributed to its success.\newline

We express our heartfelt gratitude to our colleagues, Shri.N S S Kamalesh and Shri. Sai Sadashiva, from Sri Sathya Sai Institute of Higher Learning (SSSIHL), for their unwavering support and guidance. Their significant contributions have been the backbone of this project and have played a crucial role in shaping its success.\newline

\section*{\textbf{Dedication}}
The authors from SSSIHL dedicate this paper to
the founder chancellor of SSSIHL, Bhagawan Sri Sathya Sai
Baba. The corresponding author also dedicates this paper to
his loving elder brother D. A. C. Prakash who still lives in
his heart and the first author also dedicates this paper to his loving parents Murali krishna and Sai Lakshmi.

\end{document}